\documentclass[10pt,a4paper]{article}
\usepackage[hyperref]{acl2021}
\usepackage{times}
\usepackage{latexsym}

\usepackage{microtype}
\usepackage{graphicx}

\aclfinalcopy 

\title{Training Conversational Agents with Generative Conversational Networks}

\author{
  Yen-Ting Lin, Alexandros Papangelis, Seokhwan Kim, Dilek Hakkani-T\"ur \\
  \texttt{\{ytl, papangea, seokhwk, hakkanit\}@amazon.com} \\
  }

\begin{document}
\maketitle
\begin{abstract}
Rich, open-domain textual data available on the web resulted in great advancements for language processing. However, while that data may be suitable for language processing tasks, they are mostly non-conversational, lacking many phenomena that appear in human interactions and this is one of the reasons why we still have many unsolved challenges in conversational AI. In this work, we attempt to address this by using Generative Conversational Networks to automatically generate data and train social conversational agents. We evaluate our approach on TopicalChat with automatic metrics and human evaluators, showing that with 10\% of seed data it performs close to the baseline that uses 100\% of the data.
\end{abstract}

\section{Introduction}
Conversational Artificial Intelligence (AI) has progressed a lot in the recent past, partly due to advances in large pre-trained language models (LM) and partly due to commercial conversational agents (Alexa, Siri, Cortana, Google Assistant, and others). It is evident, however, that many challenges still remain, such as handling idioms, humour, expressing empathy, building and maintaining rapport, and so on. One big factor for this is the lack of large and rich conversational data that include these complex aspects of human communication. While the research community is making great efforts in collecting such data, these are still small compared to the amount of data needed to train deep networks. 
In this work, we take a first step into automatically generating conversational data using Generative Conversational Networks (GCN) \cite{papangelis2021generative}, hoping that in the future we will be able to generate many of the aforementioned phenomena. GCN learns how to generate labelled, diverse, and targeted data that are optimised with Reinforcement Learning (RL) to maximise task performance (more details in the following section). 

There are many recent studies on data augmentation, but most of them are geared towards language processing tasks rather than training conversational agents. Due to lack of space we only mention a few. PROTODA \cite{kumar2021protoda} uses prototypical networks to augment data for intent classification while GenSF \cite{mehri2021gensf} uses DialoGPT \cite{zhang2020dialogpt} for zero-shot slot tagging; DINO \cite{schick2021generating} uses pre-trained language models (PLM) to generate data for semantic textual similarity; \cite{campagna2020zero} focus on zero-shot dialogue state tracking and use an abstract dialogue model to generate data and improve performance. Soloist \cite{peng2020soloist} uses a PLM fine-tuned on large dialogue corpora and is designed for transactional (goal-oriented) dialogues. \cite{mohapatra2020simulated} use PLM to train user simulators from crowd-generated conversations and their instructions, for transactional conversations. \cite{lin2021domain} train domain-independent user simulators for transactional dialogues. 

\begin{table*}[!ht]
\centering
\small
\caption{Automatic and human evaluation results. Human evaluators rate responses with the same conversation context on a scale of 1 to 5. BScore stands for BERTScore. Bold indicates statistical significance (t-test assuming unequal variance).}
\label{tb:human_eval}
\begin{tabular}{l|ccc|cccc}
& \multicolumn{3}{c|}{Automatic Evaluation} & \multicolumn{4}{c}{Human Evaluation} \\
Model & BLEU & Rouge(1/2/L) & BScore & Engaging. & Fluency & Relevance & Overall \\
\hline
Ground Truth & - & - & - & 3.85 $\pm$ 1.04 & 4.55 $\pm$ 0.85 & 3.77 $\pm$ 1.05 & 4.06 $\pm$ 1.04 \\
BART-100 & 3.1 & 20.3/6.1/17.8 & 0.861 &  3.80 $\pm$ 1.06 & 4.58 $\pm$ 0.85 & 3.68 $\pm$ 1.10 & 4.02 $\pm$ 1.08 \\
\hline
GCN Learner & 1.3 & 15.8/2.7/13.6 & 0.851 &  \textbf{3.79 $\pm$ 1.06} & 4.49 $\pm$ 0.95 & 3.58 $\pm$ 1.14 & \textbf{3.96 $\pm$ 1.12} \\
BART-10 &\textbf{2.0} & \textbf{18.5/4.2/16.0} & {\bf 0.858} &  3.63 $\pm$ 1.04 & 4.50 $\pm$ 0.92 & 3.62 $\pm$ 1.12 & 3.92 $\pm$ 1.11 \\
GCN-RL Learner & 1.1 & 15.0/2.1/12.6 & 0.850 &  3.70 $\pm$ 1.10 & 4.47 $\pm$ 0.99 & 3.47 $\pm$ 1.17 & 3.88 $\pm$ 1.17 \\
\end{tabular}
\end{table*}

\section{Modeling Multi-Turn Conversations}

\subsection{Generative Conversational Networks}
GCN \cite{papangelis2021generative} consist of two models in a meta-learning architecture: a data generator and a learner. The data generator creates a labelled dataset that is used to train the learner (conversational agent in our case) in a supervised fashion. The learner is then evaluated on an external validation set and its performance is used as a proxy of the quality of the dataset. This quality measure is used as a reward in a RL setup that trains the generator. This way, the generator learns to create data that lead the learner to perform well on the validation set. Both models can be pre-trained with seed data, if available, and paired with reward estimation, GCN can be used for continuous learning from user feedback. This approach has been proven to work well for intent detection and slot tagging for goal-oriented conversations and we here apply it to train social conversational agents.

\subsection{Generator training} 
We use DialoGPT \cite{zhang2020dialogpt} to frame the multi-turn conversation generation as autoregressive language modeling.
We define a multi-turn conversation as a list of utterances: $T_1$, $T_2$, $...$, $T_N$ where $T_i$ is the utterance in turn $i$, and $N$ is the number of turns in the conversation.
Each utterance $T$ is composed of tokens $x_1$, $x_2$, $...$, $x_n$, where $n$ is the number of tokens in the utterance. 

In our case, we use seed data $D$ to aid training, that are split into training and validation. The generator is given the first three turns $T_1$, $T_2$, and $T_3$ sampled from the seed data as a prompt and generates the remaining turns $T'_4$, $T'_5$, $...$, $T'_N$ in a self-playing setting. This is done for a number of batches, resulting in a new dataset $D'$. A new learner is then spawned and trained on this dataset and evaluated on the seed validation set.
Following \cite{ziegler2019fine} and \cite{papangelis2021generative}, we use PPO \cite{schulman2017proximal} to train the generator using the learner's performance on the validation set as a reward, and iterate the data generation - learner training process until performance does not change significantly (within a small tolerance).
We use a weighted sum of BLEU \cite{papineni-etal-2002-bleu}, Rouge-1 \cite{lin-2004-rouge}, and BERTScore \cite{bert-score} as the reward for the generator RL training.\footnote{The weights are used to normalize the metrics to a scale of 0 to 1 and determined empirically: $0.1$, $0.01$, $0.95$, for BLEU, Rouge-1 and BERTScore, respectively.}
After the training phase, we create a final dataset $D'$ and use that and the seed training data $D$ to train a final learner that is used for evaluation.

\section{Evaluation}
\begin{figure}
    \centering
    \includegraphics[width=1.0\linewidth]{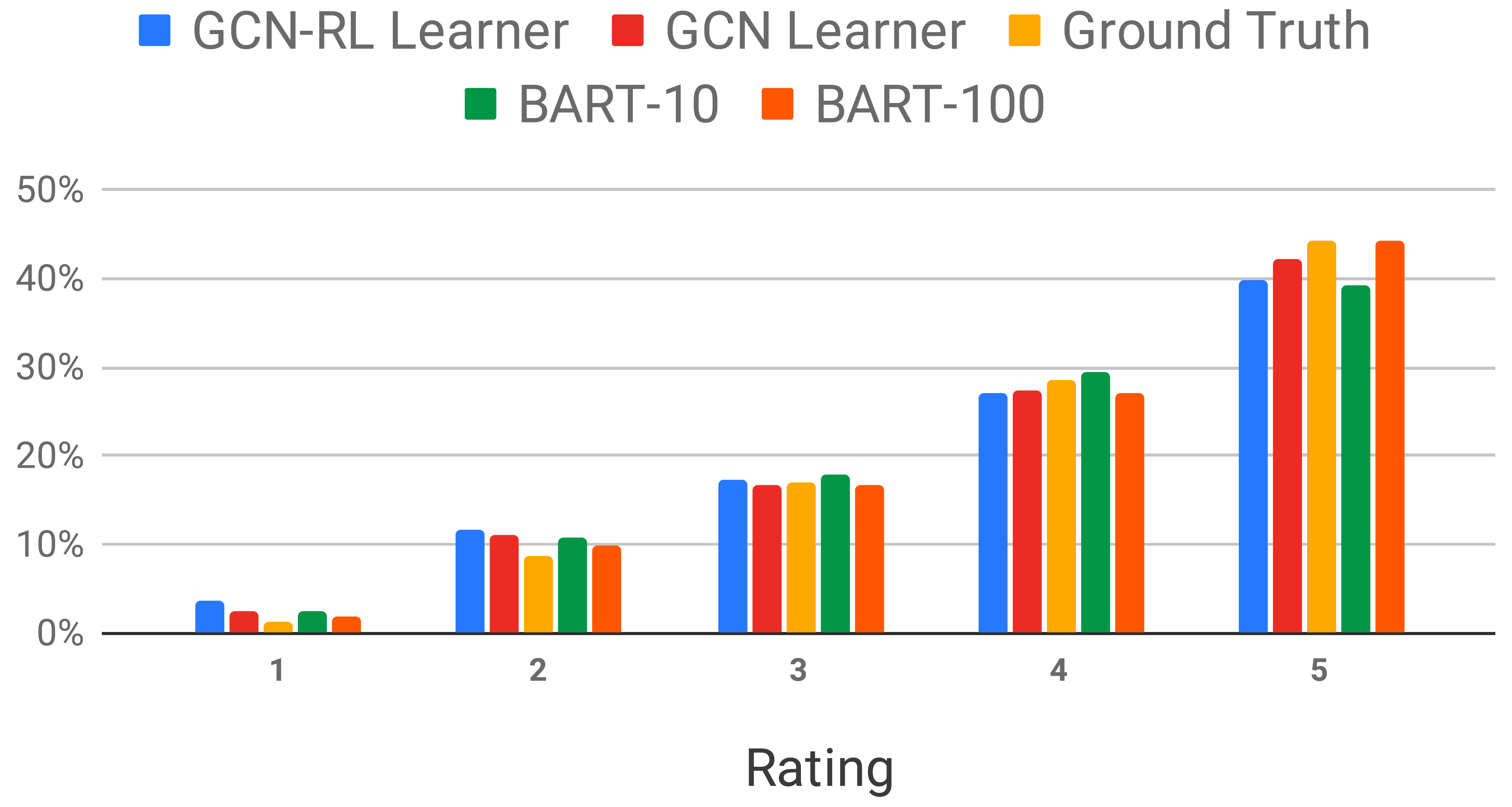}
    \caption{Distribution of ratings from human evaluators}
    \label{fig:rating_dist}
    \vspace{-4mm}
\end{figure}

We use TopicalChat \cite{gopalakrishnan2019topical} in our experiments, sampling 10\% of it as seed for GCN. 
We use DialoGPT \cite{zhang2020dialogpt} and BART \cite{lewisbart} as initial weights for the generator and the learner, respectively.
As mentioned in the previous section, our GCN learner was trained on both seed data and synthetic data, created by the GCN generator.
We compare the performance of the GCN learner against the ground truth and against three baselines: BART trained with the same seed data (BART-10), BART trained with the complete training set of TopicalChat (BART-100), and a GCN learner trained on seed and synthetic data but without updating the generator via RL (GCN-RL). \\
{\bf Automatic evaluation.} We calculate the reference-based automatic evaluation metrics on the TopicalChat test set (frequent) (Table \ref{tb:human_eval}, left).
The learner agent trained on the seed and synthetic data by the generator performs lower on most metrics.
However, due to the intrinsic one-to-many property of conversation, reference-based metrics may not correlate well with human judgments; our generated conversation may be appropriate for the dialogue context but different from the reference responses. For this reason, we also conduct human evaluation. \\
{\bf Human evaluation.} We conduct a study were human evaluators rate the output of the GCN learner, the baselines, and the ground truth. Specifically, we rate how engaging, fluent, and relevant each response from each model is, on a scale from 1 to 5. We generate 1000 samples for each condition using the same context and make sure we have 3 ratings per sample per condition. Table \ref{tb:human_eval} (right) shows the results of the evaluation, where we see that the GCN learner outperforms the BART-10 and GCN-RL models and is close to BART-100's performance. All models are outperformed by the ground truth, which may be due to the size of our models or the number of training iterations. Figure \ref{fig:rating_dist} shows the distribution of ratings for all conditions.


\section{Discussion}
We present GCN for multi-turn conversations, where our agent can learn to generate the data it needs for social conversations. The lower results in the reference-based metrics (Table \ref{tb:human_eval}, left) indicate that GCN generates new data; paired with the good scores in the human evaluation (Table \ref{tb:human_eval}, right), we see that the generated data is also useful for the task at hand, a result in line with the results on intent detection and slot tagging in \cite{papangelis2021generative}. As future work, we are conducting thorough analyses and evaluations of the generated data, exploring more generator the learner architectures, and working towards knowledge-grounded conversations.

\bibliographystyle{acl_natbib}
\bibliography{anthology,acl2021}


\end{document}